\documentclass[runningheads]{llncs}
\usepackage[T1]{fontenc}
\usepackage[compress,square,numbers,round]{natbib}
\usepackage{custom}
\usepackage{graphicx}

\makeatletter
\renewcommand{\section}{\@startsection{section}{1}{\z@}%
  {-10pt plus -2pt minus -2pt}%
  {4pt plus 2pt minus 2pt}%
  {\normalfont\normalsize\bfseries}}
\renewcommand{\subsection}{\@startsection{subsection}{2}{\z@}%
  {-9pt plus -2pt minus -2pt}%
  {2pt plus 1pt minus 1pt}%
  {\normalfont\normalsize\bfseries}}
\renewcommand{\subsubsection}{\@startsection{subsubsection}{3}{\z@}%
  {-8pt plus -1pt minus -1pt}%
  {1pt plus 1pt minus 1pt}%
  {\normalfont\normalsize\bfseries}}
\makeatother

\begin{document}
\title{Prospective Learning in Retrospect}
%
%

\author{
Yuxin Bai\inst{1, \thanks{Equal contribution}}\orcidID{0009-0007-0689-7487} \and
Cecelia Shuai\inst{1, \footnotemark[1]}\orcidID{0009-0003-9882-7566} \and
Ashwin De Silva\inst{1, \footnotemark[1]}\orcidID{0000-0002-6406-7090}  
\and
Siyu Yu\inst{1}\orcidID{0009-0001-1459-2567} \and 
Pratik Chaudhari \inst{2}\orcidID{0000-0003-4590-1956} \and  
Joshua T. Vogelstein \inst{1}\orcidID{0000-0003-2487-6237}}


%
%
\authorrunning{Y. Bai et al.}  

\institute{Johns Hopkins University \and
University of Pennsylvania \\
\email{\{ybai31, xshuai3, ldesilv2, jovo\}@jhu.edu, pratikac@seas.upenn.edu}}
\maketitle              

\begin{abstract}

In most real-world applications of artificial intelligence, the distributions of the data and the goals of the learners tend to change over time. The Probably Approximately Correct (PAC) learning framework, which underpins most machine learning algorithms, fails to account for dynamic data distributions and evolving objectives, often resulting in suboptimal performance. Prospective learning is a recently introduced mathematical framework that overcomes some of these limitations. We build on this framework to present preliminary results that  improve the algorithm and numerical results, and extend prospective learning to sequential decision-making scenarios, specifically foraging. Code is available at: \url{https://github.com/neurodata/prolearn2}.


\end{abstract}

\section{Introduction}

Learning involves updating decision rules or policies, based on past experiences, to improve future performance. The Probably Approximately Correct (PAC) learning~\citep{vapnik1991principles, Valiant2013-gp} framework has led to the development of learning algorithms that provably minimize the risk (expected loss) over unseen future samples during inference. When proving such guarantees, PAC learning assumes that data is independently and identically (\textit{iid}) distributed according to a fixed distribution at training and inference time.

While this assumption has been useful, it is rarely held true in practice. In fact, the future is more likely to be different from the past as distributions of data and goals of the learner may change over time. Therefore, the true hypothesis can be time-variant, and classical PAC learning does not address this situation. Although sub-disciplines such as transfer learning \citep{ben2010theory}, continual/lifelong learning \citep{thrun1998lifelong,vogelstein2020simple,ramesh2021model, antoniou2020defining}, online learning \citep{shalev2012online}, meta-learning \citep{finn2017model,
maurer2005algorithmic}, sequential decision-making \citep{Ghosh1991-kp, Cesa-Bianchi2006-tv}, forecasting \citep{petropoulos2022forecasting}, reinforcement learning \citep{sutton1998reinforcement, chen2022you, kumar2023continual, levine2020offline}, out-of-distribution generalization \citep{de2023value} have  introduced attractive solutions that retrospectively adapt to distributions that change over time, they often fail to anticipate and generalize to future even when data evolves in simple but predictable ways as shown in~\citep{NEURIPS2024_de85d3cf, de2023prospective}.

Prospective learning~\citep{NEURIPS2024_de85d3cf, de2023prospective} is a recently developed mathematical framework to bridge this gap. Instead of data arising from a fixed distribution, prospective learning assumes that data is drawn from a stochastic process, that the loss considers the future, and that the optimal hypothesis may change over time. A prospective learner uses samples received up to some time $t \in \naturals$ to output an infinite sequence of predictors, which it uses for making predictions on data at all future times $t' > t$. An exhaustive comparison between prospective learning and related sub-fields of machine learning literature is provided in~\citet{NEURIPS2024_de85d3cf}.

We build on the prospective learning framework and introduce several preliminary results. The rest of the paper is organized as follows; \Cref{s:preliminaries} provides a concise summary of the prospective learning framework, \Cref{s:observations-mlp} presents several empirical observations on deep learning-based prospective learners, \Cref{s:forest} introduces a decision-tree based prospective learner and demonstrates its performance, and finally, \Cref{s:foraging} introduces prospective foraging, showing that the prospective learning framework extends beyond supervised learning and performs competitively with reinforcement learning in preliminary results.

\section{Preliminaries}
\label{s:preliminaries}

Let input and output be denoted by $x_t \in \mc{X}$ and $y_t \in \mc{Y}$, respectively. Let $z_t = (x_t, y_t)$. We will find it useful to denote the observed data, $z_{\leq t} = \{z_1, \dots, z_t\}$, and the unobserved data, $z_{>t}$. In contrast to PAC learning, $t$  is not just a dummy variable, but rather, indexes time. We therefore define the data triple $(x_t,y_t,t)$, and augment the input space to include  the time of the input, $\mc{X} \leftarrow \mc{X} \times \mc{T}$. Consider an infinite sequence of hypotheses $h \equiv (h_1,\dots,h_t,h_{t+1},\dots)$ where $h_t : \mc{X} \to \mc{Y}$. The hypothesis class $\mc{H}$ is the space that contains such sequences. With a slight abuse of notation, we will refer to a sequence of hypotheses $h$ as a \textit{hypothesis}, where each element of this sequence $h_t: \XX \mapsto \YY$. \footnote{One could also think of prospective learning as using a single time-varying hypothesis $h: \naturals \times \XX \mapsto \YY$, i.e., the hypothesis takes both time and the datum as input to make a prediction.}



\textbf{Loss}
The instantaneous loss, $\ell( h(x),y)$, is a map from $\mc{Y} \times \mc{Y}$ to $\mathbb{R}$.  In all real-world problems we care about the integrated future loss. 
Let $w(i)$ be a non-increasing non-negative weighting function that sums to one, that is $\sum_i w(i) = 1$ and $0 \leq w(i) \leq 1\, \forall i$.  We thus 
define prospective loss as 
\beq{
\bar{\ell}(h, z_{>t}) = \sum_{s>t} w(s-t) \ell(h(x_s),y_s)
\label{eq:ell_bar}
}
which is the weighted cumulative loss over all the future data.
\footnote{If we let $w(i)$ be a constant function of $i$,  we recover the classical loss in PAC learning.}

Taking the expectation of \cref{eq:ell_bar} over the future conditioned on the observed data $z_{\leq t}$, we arrive at the prospective risk at time $t$,
\beq{
    R_t(h)
    = \E \sbr{\bar \ell_t(h,Z) \mid z_{\leq  t}}
    = \int \bar \ell_t(h,Z) \dd{\P_{Z \mid z_{\leq t}}}.
    \label{eq:prospective_risk}
}
\footnote{This is a slight abuse of notation, because previously $\ell$ mapped from a hypothesis and data corpus, but now we are saying that it maps from a hypothesis and a collection of random variables. 
We have used the shorthand  $\E[Y \mid x]$ for  $\E[Y \mid X = x]$. 
Note that $\mathbb{E}[Z | z_{\leq t}]$ is equivalent to $\mathbb{E}[Z_{>t} | z_{\leq t}]$ because the past $z$'s are given. 
}

We consider a family of stochastic processes $\mc{Z}$ to be strongly prospective-learnable if there exists a learner that likely returns an approximately optimal hypothesis (with a risk close to the Bayes risk $R_t^\ast$) after observing enough past data $z_{\leq t}$ from any process $Z \in \mc{Z}$. Theorem 1 from~\citet{NEURIPS2024_de85d3cf} guarantees that under certain general assumptions, a slightly modified Empirical Risk Minimizer (ERM)
a learner returning the hypothesis
\beq{
\hat{h} =
\argmin_{h \in \mc{H}_t}  \sum_{t'>0}^t \bar{\ell}(h,z_{>t'}) = \argmin_{h \in \mc{H}_t} \sum_{t' > 0 }^t \sum_{s> t'} w(s-t') \ell(h_{t'}(x_s),y_s).
\label{eq:prospective_erm}
}
is a strong prospective learner for a finite family of stochastic processes under certain assumptions. We refer to this learner as Prospective ERM. 

To implement prospective ERM in practice, one may modify a predictor (e.g. a multi-layer perceptron) to take $(\varphi(s), x_s)$ as the input and train it to predict the label $y_s$, where $\varphi(s)$ is a suitable embedding on the time $s$. We refer to this predictor as \textit{Prospective-MLP}. Inspired by the positional encoding of the Transformer~\cite{vaswani2017attention}, \citet{NEURIPS2024_de85d3cf} have used the time-embedding,
\beq{
\varphi_{f}(t) = (\sin(\omega_1 t), \dots, \sin(\omega_{d/2} t), \cos(\omega_1 t), \dots, \cos(\omega_{d/2} t)),
\label{eq:fourier-embed}
}
where \(\omega_i = \pi / i\) for \(i = 1, \dots, d/2\). It is shown that the prospective-MLP converges to the Bayes risk on certain stochastic processes over synthetic and image data.

\subsubsection{Prospective Learning Scenarios}
Depending on the nature of the stochastic process, one can consider 4 scenarios of prospective learning; (1) Data is independent and identically distributed, (2) Data is independent but not identically distributed, (3) Data is neither independent nor identically distributed (e.g. Markov Processes), and (4) Future depends on the current prediction (Stochastic Decision Processes). Scenario 1 puts us back in the PAC learning setting. Scenario 4, on the other hand, is arguably a special case of Scenario 3 and has implications for reinforcement learning and control. As we will introduce in~\cref{s:foraging}, prospective foraging is closely related to this scenario.

\subsubsection{Example Stochastic Processes} \label{sss:processes}
Here we describe three stochastic processes we will consider in the experiments outlined in~\cref{s:observations-mlp,s:forest}.

\begin{figure}[htb]
  \centering
  \begin{subfigure}[c]{0.23\textwidth}
    \centering
    \includegraphics[width=\linewidth]{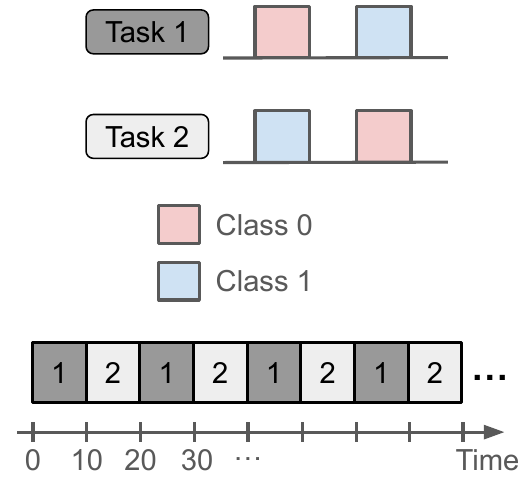}
  \end{subfigure}
  \begin{subfigure}[c]{0.23\textwidth}
    \centering
    \includegraphics[width=0.8\linewidth]{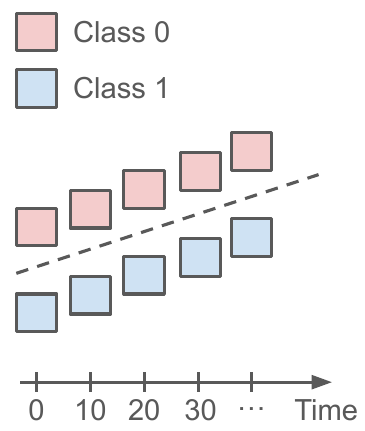}
  \end{subfigure}
  \begin{subfigure}[c]{0.45\textwidth}
    \centering
    \includegraphics[width=1.1\linewidth]{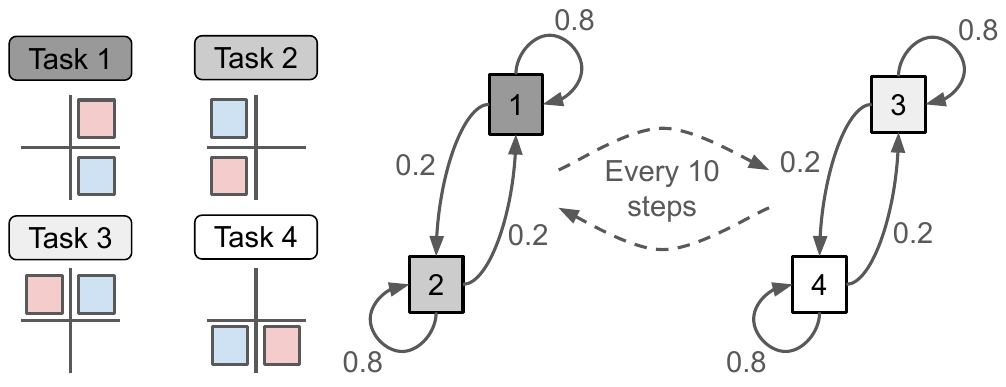}
  \end{subfigure}
  \caption{Pictorial depictions of 3 types of stochastic processes considered in our experiments. (\textbf{Left}) Periodic Process, (\textbf{Middle}) Linear process, and (\textbf{Right}) Hierarchical hidden Markov Process. The periodic and linear processes belong to scenario 2 whereas the hierarchical hidden Markov process is an instance of scenario 3. }
  \label{fig:processes}
  \vspace{-0.7em}
\end{figure}

First, consider two binary classification distributions (``tasks'') (see~\cref{fig:processes} (Left)). The inputs for both tasks are drawn from a uniform distribution on the set $[-2, -1] \cup [1, 2]$. Ground-truth labels correspond to the sign of the input for Task 1, and the negative of the sign of the input for Task 2. The process switches the two tasks every $10$ time steps, resembling a reversal learning problem. We refer to this as the ``periodic'' process.

The second is a stochastic process whose marginal distribution at time $t$ is defined as follows: The input $x_t$ is drawn from a uniform distribution over $[\epsilon t+10, \epsilon t+11] \cup [\epsilon t-10, \epsilon t-11]$ and its label $y_t$ is $0$ if $x_t > t$ and $1$ otherwise. In other words, it is a process where the task is hiking up a slope with a small gradient $\epsilon$ (see~\cref{fig:processes} (Middle)). This process yields an infinite number of tasks, in contrast to the first one, where there are finite (two) tasks. Thus, we refer to it as the ``infinite task process''. 

The third and final process includes four tasks that are created using 2-dimensional inputs as shown in~\cref{fig:processes} (Right). After every 10 time-steps, a different Markov chain would govern the transitions among tasks (one Markov chain for tasks 1 and 2, and another for tasks 3 and 4 as illustrated in the figure). Therefore, the data distribution is effectively distributed according to the hierarchical hidden Markov model. We refer to it as the ''dependent structured task process''. 

\vspace{-3pt}
\subsubsection{Relationship between Continual and Prospective Learning} 
Although continual and prospective learning both involve learning over sequences of tasks, they differ fundamentally in their objectives and assumptions. As formalized in~\cref{s:preliminaries}, the goal of a prospective learner is to perform well on future tasks. In contrast, the objective in continual learning, though typically less formally defined, is to maintain good performance on previously seen tasks and avoid catastrophic forgetting. Continual learning often assumes a task-aware setting, where the learner receives a batch of data per task along with the task identity. Prospective learning, by contrast, does not assume access to such task labels or boundaries. A notable exception is task-agnostic online continual learning~\citep{zeno2021task}, which operates under similar assumptions to prospective learning. However,~\citet{NEURIPS2024_de85d3cf} shows that such methods still fail to improve upon chance-level prospective risk when learning the periodic process above. Furthermore, continual learning benchmarks typically involve sequences of unrelated tasks without any predictable structure, whereas prospective learning is meaningful when the tasks evolve over time in a predictable manner. To address this gap, we introduce simple yet representative benchmarks designed to assess the ability of learners to generalize to future tasks. For additional experiments and a deeper comparison of prospective learning with related paradigms, we refer the reader to~\citet{NEURIPS2024_de85d3cf}.

\subsubsection{Training and Evaluating Learners}
The next two sections present experiments including Prospective-MLPs and Prospective-Trees, and a time-agnostic Follow-the-Leader (FTL) baseline that minimizes empirical risk over all past data without incorporating time as an input. When training and evaluating these learners, we roughly follow the steps detailed in the Section 6 of~\citet{NEURIPS2024_de85d3cf}.

\section{Several Empirical Observations on Prospective-MLPs}
\label{s:observations-mlp}

\subsubsection{Prospective-MLP prevails under heterogeneous sampling.} 
\begin{figure}[ht]
  \centering
  \includegraphics[width=0.7\linewidth]{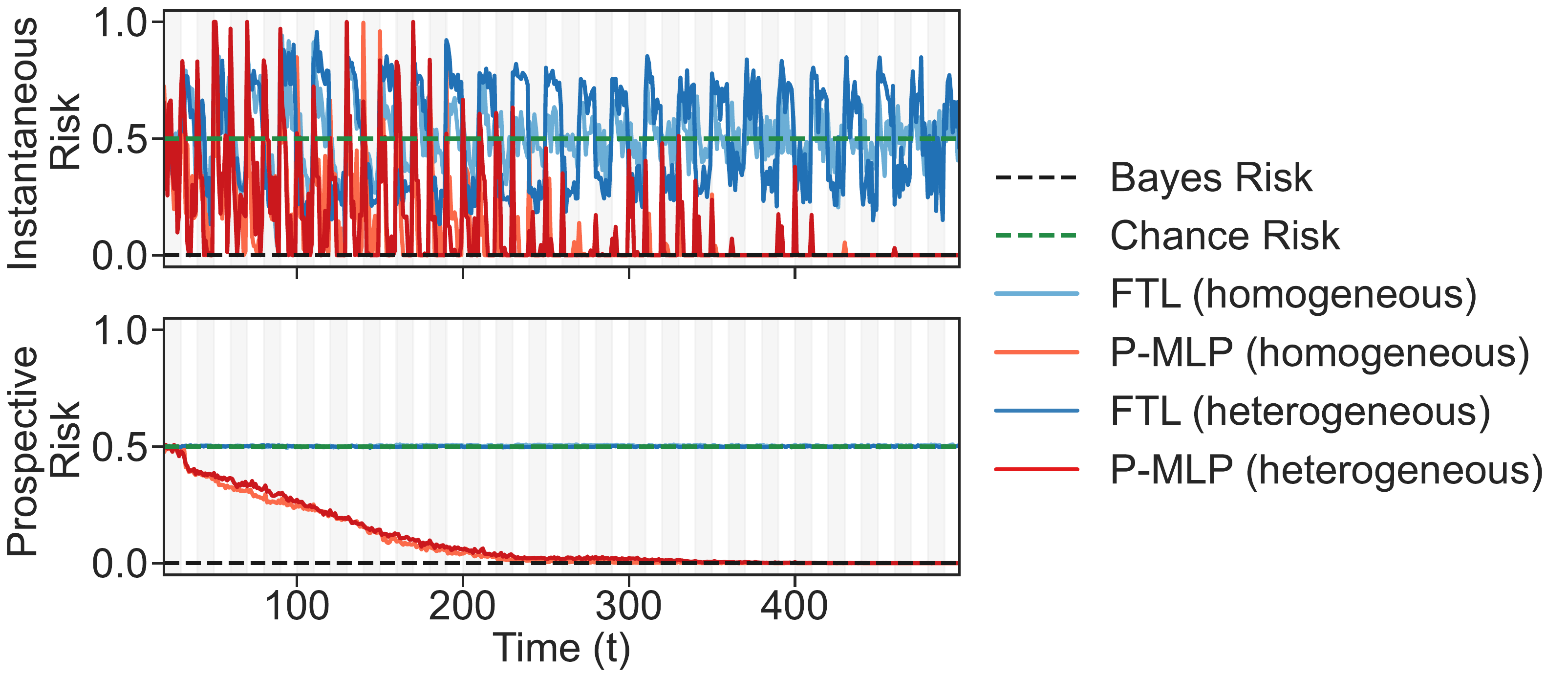}
  \caption{Instantaneous (\textbf{top}) and prospective (\textbf{bottom}) risks of Follow-the-Leader (FTL, blue) and Prospective-MLP (P-MLP, red) trained on homogeneously (lighter shade) and heterogeneously (darker shade) sampled data from the periodic process. Homogeneous sampling is where you get exactly one sample each time step. In heterogeneous sampling, there can be missing samples and/or multiple samples available per time step.}
  \label{fig:effect_of_sparsity}
  \vspace{-0.7em}
\end{figure}

The experiments in~\citet{NEURIPS2024_de85d3cf} assume homogeneous past data where exactly one sample is received at each time step. This assumption overlooks more realistic scenarios where samples may be missing or multiple samples may be available per time step. To model the heterogeneity of sampling, we assume that the number of samples received from the process at each time step is distributed according to a Poisson distribution with $\lambda = 1$. We train Follow-the-Leader (FTL) and prospective-MLP on data collected this way and compare them against their counterparts trained on homogeneous data (see~\cref{fig:effect_of_sparsity}). It is evident that prospective-MLP manages to secure a good prospective risk regardless of how the data is sampled.

\subsubsection{Prospective-MLP prevails in infinite task scenarios.} Experiments in~\citet{NEURIPS2024_de85d3cf} are mostly based on stochastic processes that include several tasks that periodically switch between each other. On such processes, it has been shown that the Prospective-MLPs equipped with the time-embedding defined by~\cref{eq:fourier-embed} are able to achieve a low prospective risk and generalize over the future. It is intuitive that a time-embedding comprised of Fourier basis functions is appropriate for a periodic process assuming that it contains the function with the true switching frequency. 

However, aside from capturing periodic patterns, the utility of a Fourier-based time embedding is likely to be limited. To demonstrate this, we consider the linear process (see~\cref{sss:processes} and~\cref{fig:processes} (Middle)). There, we get a new task at each time and the task evolves according to a linear trend in time. The prospective learner must exploit this trend in order to generalize over the future. To illustrate the effect the choice of time-embedding has on the learner, we train two Prospective-MLPs, one with the Fourier embedding from~\cref{eq:fourier-embed} and the other with a time-embedding based on monomial basis functions given by $\varphi_{m}(t) = (t, t^2, t^3, \dots, t^d)$. We repeat the same routine with the periodic process (see~\cref{sss:processes} and~\cref{fig:processes} (Left)) as well.

\begin{figure}[htb]
\vspace{-0.3em}
  \centering
  \includegraphics[width=\linewidth]
  {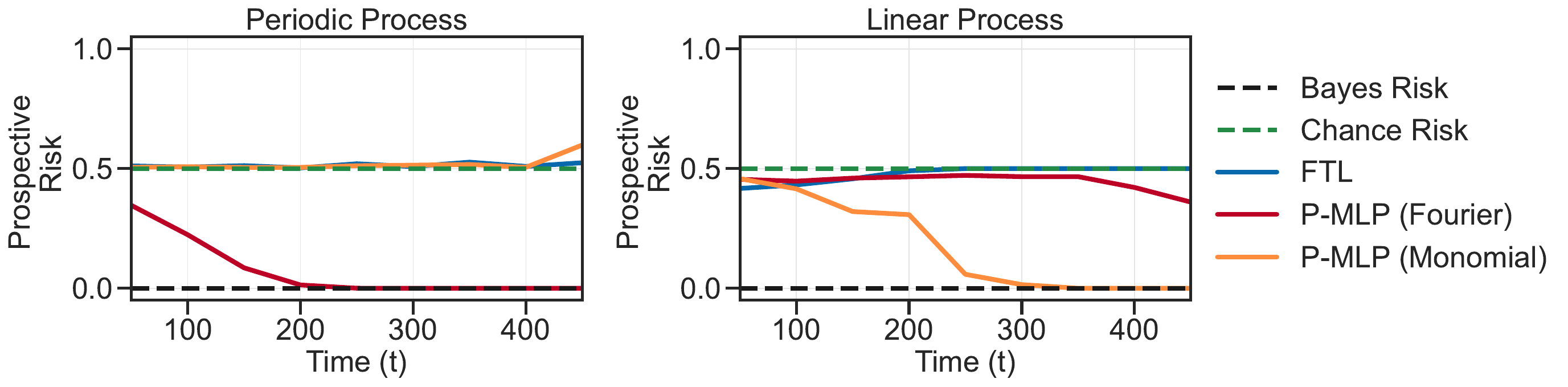}
  \vspace{-1em}
  \caption{Prospective risk of Follow-the-Leader (FTL), and Prospective-MLP with Fourier embeddings, and Prospective-MLP with monomial embeddings on periodic (Right top) and linear (Right bottom) processes. Prospective-MLP with Fourier embeddings performs best on the periodic process, whereas the variant with monomial embeddings achieves the best performance on the linear process.}
  \label{fig:effect_of_time_embedding}
  \vspace{-0.5em}
\end{figure}

As expected, the Prospective-MLP with the monomial embedding outperforms other learners trained on the samples from the linear process with an infinite number of tasks. However, it fails to perform well on the periodic process, where the Fourier embedding is more appropriate. 
The key takeaways from this experiment is that prospective learning can perform well even when there are an infinite number of tasks, but to do so, it must leverage an appropriate time-embedding for the underlying process. 

\subsubsection{Prospective-MLPs can be trained in a streaming or online manner.} 
\begin{figure}[htb]
\vspace{-1em}
    \centering
    \includegraphics[width=0.6\linewidth]{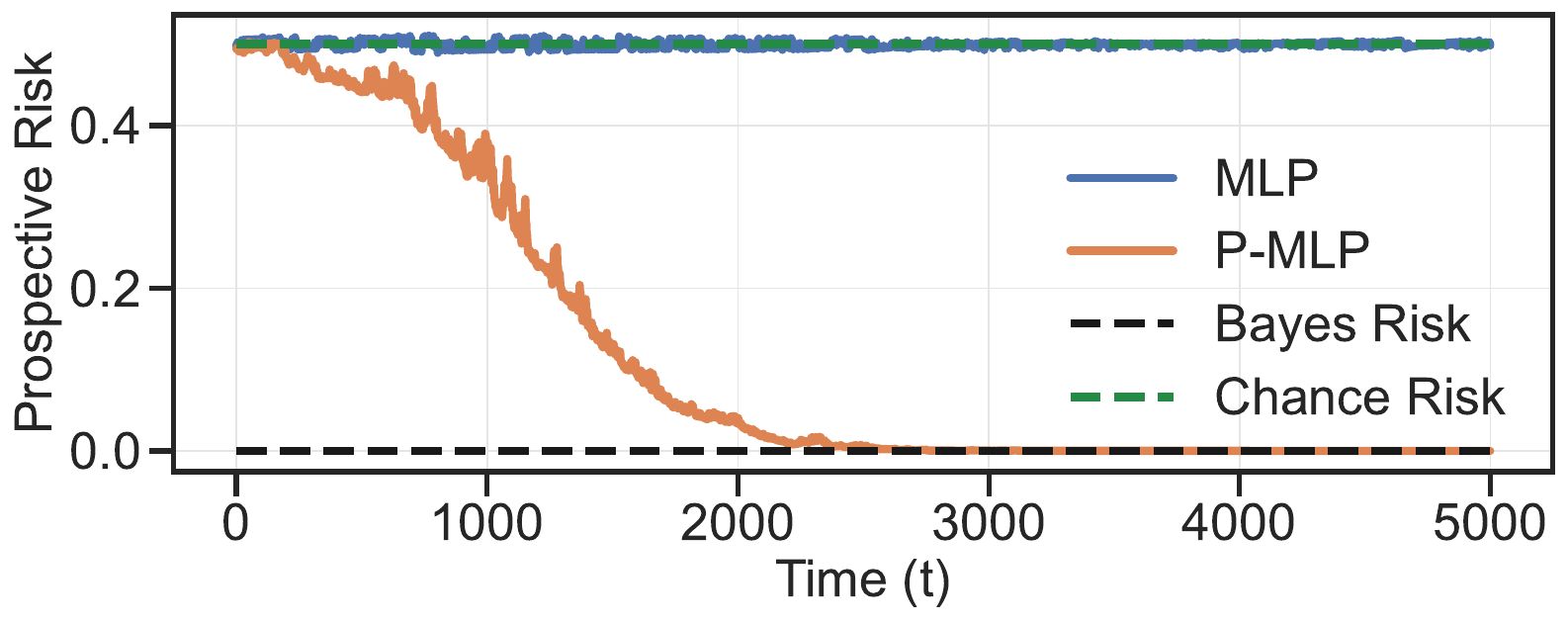}
    \vspace{-0.5em}
    \caption{Prospective risk of the learners that are trained in an online manner on data from the periodic process.}
    \label{fig:online_training}
    \vspace{-0.7em}
\end{figure}

So far, we have considered scenarios where Prospective-MLPs are trained in an offline or batched setting, using a fixed dataset of past samples drawn from the process. This requires the learner to have access to a memory where it may store the dataset used for training. When there are constraints on the memory allowed for the learner, batched-learning is no longer feasible. Here, we consider an extreme but realistic setting, where the learner will see the sample drawn from the process at time $t$ only once. Therefore, the learner is expected to perform a parameter update after observing each new sample. In~\cref{fig:online_training}, we plot the risk of a Prospective-MLP that was trained in this manner over the data sampled from the periodic process. Notice that the batch-trained Prospective-MLP converges to the optimal risk within approximately 250 samples (see~\cref{fig:effect_of_sparsity} (Bottom)), whereas its online-trained counterpart requires nearly 10 times as many samples to reach the same level of performance. This is expected as the model is exposed to the same training datum more than once during batched-training. 

\section{ Prospective Forests}
\label{s:forest}

\subsection{Motivation and background}

Decision forests, including Random Forests (RF) and Gradient Boosted trees (GBTs) continue to empirically outperform deep learning methods on tabular and vector-valued data \cite{Fernandez-Delgado2014-qu,Grinsztajn2022-uy} while offering superior interpretability. However, most existing results are for problems under the assumptions of the PAC framework \cite{biau_08_consisitency_rf}. In addition to the strong theoretical guarantees, including universal consistency \cite{biau_08_consisitency_rf,Scornet_2015_consisitency_rf,klusowski2023largescalepredictiondecision}, decision forests can be efficiently trained in parallel or sequentially \cite{CART1984}. 
Motivated by these strengths, we extend decision forests to the prospective learning regime. Our preliminary results indicate that Prospective Decision Trees perform comparably to the deep learning-based Prospective-MLP. 

Conventional CART (Classification and Regression Trees), as defined in \citet{CART1984}, is a greedy algorithm that builds a hierarchical structure through recursive binary splitting. We define a prospective variant of CART in the following.

\begin{definition}[\textbf{Prospective CART}]
Consider $\mathcal{Z}$ to be a finite family of stochastic processes. Suppose there is an increasing sequence of hypothesis class $\HH_1 \subseteq \HH_2 \subseteq \dots$ with each $\HH_t \subseteq (\YY^\XX)^\naturals$. $\HH^{tree}_t$  is a subset of $\HH_t$ and is the collection of all hypothesis $h \in \HH^{tree}_t$ returned by decision trees regressor. We define Prospective CART as the learner minimizes the empirical risk over past data $z {\leq} t$, i.e:
\beq{
    \hat h = \argmin_{h \in \HH^{tree}_t} \sum_{t' > 0 }^t \sum_{s> t'} w(s-t') \ell(h_{t'}(x_s),y_s),
    \label{eq:PTR}
}
where $w(i)$ is non-increasing non-negative weighting function defined in Section \ref{s:preliminaries}
\end{definition}
Naturally, random forest is a randomized ensemble of decision trees; analogously, aggregating prospective trees yields a prospective forest. However, another ensemble technique, GBTs often outperform RF \cite{friedman2001greedy,hastie2009elements} in certain PAC settings. Therefore, we also introduce a prospective version of GBTs. 


\begin{definition}[\textbf{Prospective Gradient Boosted Trees}]
Moreover, based on the definition of prospective forests, we defined prospective gradient boosting trees (GBTs) also as an ensemble of trees $h^B_t = \sum_{b=1}^B w^b_t h^b_t(x;\Theta^b_t)$. Unlike the prospective forests defined above, each tree grows independently. In prospective GBTs, the parameters $\Theta^b_t$ and the weights $w^b_t$ are iteratively updated by minimizing the empirical risk. This iterative process ensures that each step improves the model by reducing the residual error over the past data $z_{\leq t}$.
i.e. $$
    \hat h = \argmin_{h^B \in lin(\HH^{tree}_t)} \sum_{t' > 0 }^t \sum_{s> t'} w(s-t') \ell(h_{t'}(x_s),y_s), 
    \quad \text{subject to } h^B = \sum_{b=1}^B w^b_t h^b_t(x;\Theta^b_t)
    \label{eq:prospective_GBT},
$$
where $lin(\HH^{tree}_t)$ is the set of all linear combinations of functions in $\HH^{tree}_t$.
\end{definition}

\subsection{Preliminary results}
  
We consider data drawn from a periodic process and the Hierarchical Markov Process  described in~\cref{sss:processes} and illustrated in~\cref{fig:processes}. To each data point from these processes, we append two additional noise dimensions sampled from a standard normal distribution, ensuring the presence of both informative and noisy components.



\begin{figure}[htp]
    \centering
    \includegraphics[width=0.90\textwidth]
    {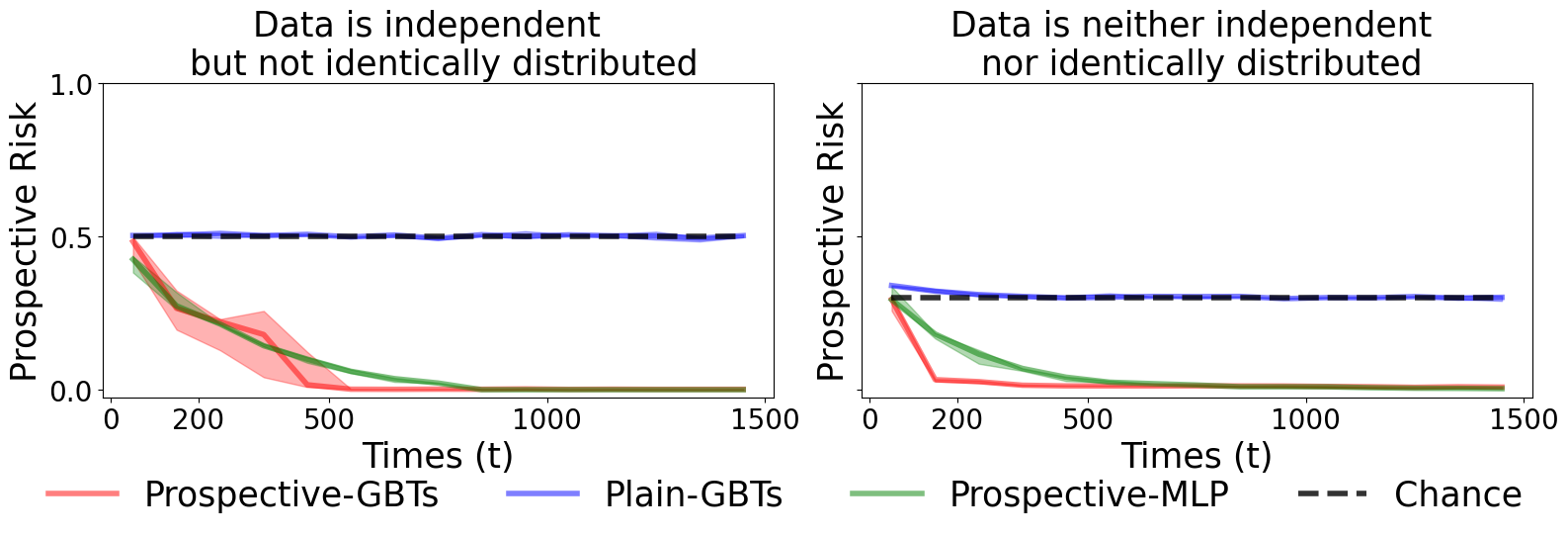}
    \caption{Prospective risk of Prospective-GBTs (red), Prospective-MLP (green) and Time-agnostic Gradient Boosted trees (Plain-GBTs, blue) across two scenarios where, (1) data is independent but not identically distributed (\textbf{Left}), and (2) data is neither independent nor identically distributed (\textbf{Right}). In both cases, the risk of Prospective-GBTs and Prospective-MLP approach the Bayes risk, with Prospective-GBTs converging faster. In contrast, the time-agnostic GBTs do not converge consistently. For comparison, the chance prospective risk is $0.5$ in the left panel and $0.3$ in the right panel.}
    \label{fig:prospective-forests}
    \vspace{-0.7em}
\end{figure}

As discussed in~\cref{s:preliminaries}, we implement Prospective-GBTs by giving it $(x_s, \varphi(s))$ as input and training it to predict the label $y_s$, where we use the time-embedding defined in~\cref{eq:fourier-embed}. In~\cref{fig:prospective-forests}, we compare the prospective risks between several learners including Prospective-GBTs and Prospective-MLP.

\vspace{-3pt}
\section{Prospective Foraging}
\label{s:foraging}

\subsection{Motivation and background}
Foraging—searching for food, water, and mates—is vital for survival and reproduction, relying on predictions of environmental fluctuations and resource availability. However, standard machine-learning and reinforcement learning (RL) methods—whether minimizing past errors or requiring extensive trial-and-error—are ill-suited to the real-time, single-lifetime risks of foraging. To bridge this gap, we introduce the prospective learning framework in which agents project into possible future states under a one-life constraint. We implement it in a simplified OpenAI Gym foraging scenario~\citep{brockman2016openai}, compare standard actor-critic RL agents~\citep{konda1999actor,shuvaev2020r} to prospectively augmented versions. Finally, we show that prospective learning framework can be extended beyond the supervised learning problem and that it outperforms an actor critic RL algorithm in a foraging task.

At each discrete time step \(t\in\mathbb{N}\) the agent observes
\(x_t=(s_t,a_{t-19:t})\in\mathcal X\)—
its current spatial location and the last \(20\) actions—and receives a
scalar reward \(y_t\in\mathcal Y\).
The only data it may inspect is the trajectory
\(z_{\le t}=\{(x_s,y_s)\}_{s=1}^t\). Standard on-policy control maximizes the generalized advantage
estimator (GAE) \(\hat A^{\text{GAE}}_t\). For notational uniformity we instead minimize the loss \(\ell(t,\hat y_t,y_t)=-\hat A^{\text{GAE}}_t\). We define a prospective forager that minimizes the sum of weighted cumulative instantaneous losses on the observed trajectory.
  
\begin{definition}[\textbf{Prospective Forager}] Consider $\mathcal{Z}$ to be a finite family of stochastic processes. Let $\mathcal H_1\subseteq\mathcal H_2\subseteq\cdots \subseteq(\mathcal Y^{\mathcal X})^{\mathbb N}$ be an expanding hypothesis class. We employ an actor–critic architecture: the critic evaluates any \(h\in\mathcal H_t\), whereas the actor can selects from \(\mathcal H^{\text{actor}}_t\subseteq\mathcal H_t\). We define Prospective Forager as the learner that minimizes the empirical risk over past data $z_{\leq} t$, i.e:
\beq{
    \hat h = \argmin_{h \in \HH^{actor}_t} \max_{u_{it} \leq m \leq t}\frac{1}{m} \sum_{s=1}^m \frac{1}{m-s+1} \sum_{r=s}^m \ell(s, h_s(x_s), y_s)
    \label{eq:actor-critic}
}
\end{definition} 
Due to the double summation in~\cref{eq:actor-critic}, more recent events are weighted more heavily when minimizing the empirical risk.

\subsection{Preliminary results} 
\subsubsection{Experiments Setting}
\begin{figure}[htp]
    \centering
    \includegraphics[width=0.85\linewidth]
    {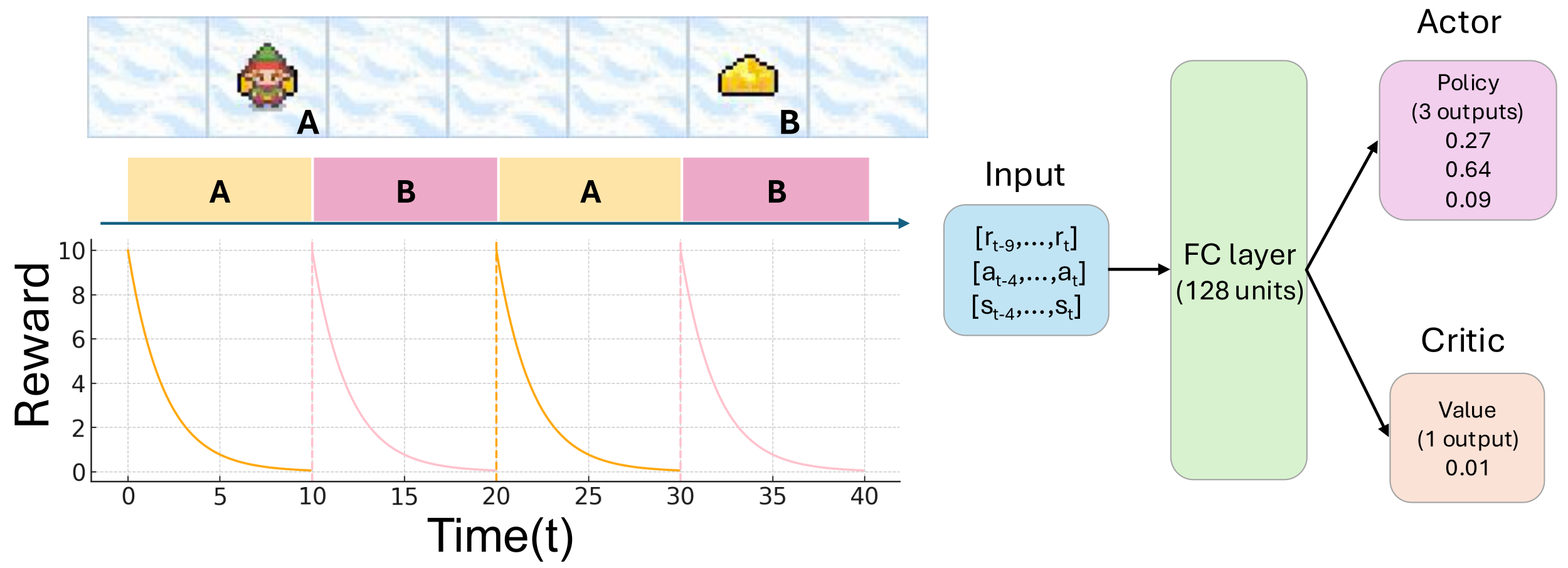}
    \caption{(\textbf{Left}) Schematics of the foraging task. Agents forage in a $1 \times 7$ linear track with two reward patches (A and B), whose reward decays in an exponential fashion. Active patches with reward availability alternates every 10 timesteps. (\textbf{Right}) The actor-critic architecture used for retrospective and prospective agents.}
    \label{fig:foraging_schematics}
    \vspace{-0.7em}
\end{figure}

An agent forages along a 1 × 7 linear track that contains two reward patches, A and B, positioned three grid spaces apart (see~\cref{fig:foraging_schematics} (Left)). Reward availability alternates between the two patches every 10 timesteps. Once the reward availability starts, the reward amount decays in an exponential fashion. A reward can be collected at time $t$ only if the agent is at the patch that is available at that moment. The agent moves one grid per timestep, and traveling between A and B takes at least three timesteps. The goal of the agent is to maximize the total amount of reward in its single lifetime, which means that there is no reset in location or time within each run. 

\vspace{-5pt}
\subsubsection{Optimal Solution}
Since the reward function is fixed, we can derive an optimal foraging strategy: the agent should leave the current patch (Patch A) before its reward is depleted and arrive at the next patch (Patch B) exactly when its reward peaks. Since no reward is available during travel, an optimal agent accepts zero immediate reward (during travel) over a low immediate reward  (by staying in Patch A), in order to maximize future reward. Hence, an agent has to prospect into the future to learn the optimal solution.


\begin{figure}[htb]
    \centering
    \includegraphics[width=0.85\linewidth]
    {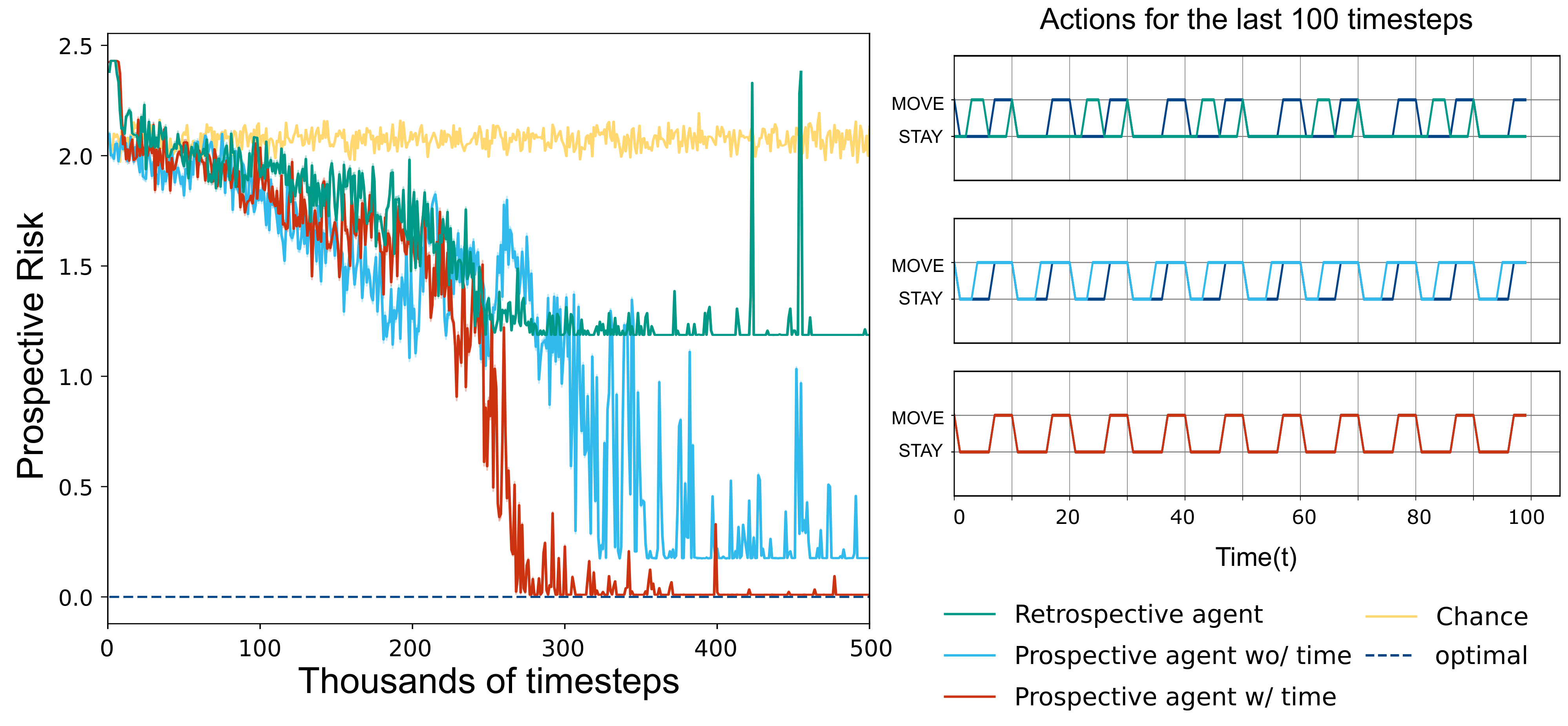}
    \caption{(\textbf{Left}) Prospective risk of prospective agents (blue and red) and retrospective agents (green) in foraging task, compared to Bayes risk (dotted blue) and chance (yellow) performance. The prospective agent without time converges to a suboptimal risk closer to Bayes risk than retrospective agent, whereas prospective agent with time converges to Bayes risk. (\textbf{Right}) Agent actions plotted for the last 100 timesteps of training. Prospective agent with time embedding (red) shows the same action plan as the optimal agent, prospective agent without time embedding (blue) leaves the patch few steps earlier than optimal, and retrospective agent does not follow the action patterns of the optimal agent.}
    \label{fig:foraging_results}
\end{figure}

\vspace{-5pt}
\subsubsection{Preliminary Result} We first implemented a standard actor-critic RL agent in the prospective foraging environment. The model architecture is described in~\cref{fig:foraging_schematics} (Right). We then implemented the prospective forager by integrating prospective risk minimization into the actor-critic agent. Finally, we added time to the prospective forager, using the same time embedding defined by~\cref{eq:fourier-embed}. At each timestep, time embedding is concatenated to the input $x_t$ in addition to agent state and action. 
\cref{fig:foraging_results} shows that the prospective agent significantly outperforms the standard actor-critic algorithm, and the inclusion of time further improves the performance to near optimal.  

\section{Conclusion}

In this work, we presented preliminary results that extend the prospective learning framework. We began by revisiting the foundational concepts and then analyzed the behavior of deep learning-based prospective learners by experimenting with heterogeneous sampling, two different choices of time-embeddings, and online-training. Next, we introduced Prospective-Trees, a nonparametric alternative that offers competitive performance against Prospective-MLPs. Finally, we proposed prospective foraging, demonstrating the framework's potential beyond supervised learning settings and highlighting its promise in sequential decision-making tasks. Collectively, these results motivate further mathematical and algorithmic exploration of prospective learning to improve learning in dynamic environments.


%
%
\bibliographystyle{unsrtnat}
\bibliography{references}


\end{document}